\documentclass[runningheads]{llncs}
\usepackage[T1]{fontenc}
\usepackage{graphicx,verbatim}
\usepackage{multirow}
\usepackage{amsmath}

\begin{document}
\title{Through the Schrödinger Bridge: Benchmarking Antemortem Image Restoration from Postmortem Autolysis to Enhance Forensic Diagnostics}

\titlerunning{Postmortem Autolysis Restoration}
%

\author{
Shuang Hao\inst{1,4} \and 
Jiacheng Yue\inst{2} \and
Yaxuan Zhao\inst{3,4} \and 
Fan Wang\inst{1,4} \and 
Jianhua Ma\inst{1,4} \and 
Erwen Huang\inst{2,*} \and 
Chunfeng Lian\inst{3,4,*}
}

\authorrunning{Hao et al.}

\institute{
Key Laboratory of Biomedical Information Engineering of Ministry of Education, School of Life Science and Technology, Xi'an Jiaotong University, Xi'an, China
\and
Faculty of Forensic Medicine, Zhongshan School of Medicine, Sun Yat-Sen University, Guangzhou, 510080, China
\and
School of Mathematics and Statistics, Xi’an Jiaotong University, Xi’an, China
\and
Research Center for Intelligent Medical Equipment and Devices (IMED), Xi’an Jiaotong University, Xi’an 710049, China
\\
\email{huangerw@mail.sysu.edu.cn, chunfeng.lian@xjtu.edu.cn}
}
  
\maketitle              
\begin{abstract}
Forensic histopathology, essential for determining cause of death and disease diagnosis, is severely impeded by postmortem autolysis, i.e., an irreversible, stochastic degradation process that distorts tissue morphology and introduces diagnostic subjectivity, thereby underscoring the value of restoring autolyzed images to a diagnostically plausible, pre-autolysis state for improving objectivity in forensic practice. This restoration task is fundamentally challenging due to the large, non-deterministic morphological changes caused by autolysis and the infeasibility of pixel-wise paired data, which invalidates assumptions underlying supervised and cycle/structure-consistent unpaired translation methods. To address this, we formalize forensic histopathology autolysis restoration as a new task: under unpaired supervision, transform postmortem images with severe autolysis into diagnostically meaningful ``antemortem'' representations. We contribute AutoPath, the first homologous yet unpaired dataset for this problem, constructed by splitting specimens into adjacent tissue blocks---one processed immediately, the other exposed to induce autolysis---yielding nearly ten thousand $10\times$ patches from 69 cases with varying liver conditions. We further frame the problem as a Schrödinger Bridge between the autolyzed and non-autolyzed distributions, offering a principled approach to modeling stochastic, severe morphological degradation. Critically, we demonstrate the misalignment of generic image-level generative metrics (e.g., FID) with diagnostic utility and propose a forensically grounded, slide-level diagnostic distribution consistency evaluation. Overall, this work establishes a reproducible benchmark (encompassing task definition, a real-world dataset, and an evaluation methodology) toward rigorous and practically meaningful progress in autolysis restoration for forensic pathology. The dataset and code will be released at \url{https://github.com/fengluo233/AutolysisBench}.

\end{abstract}

\section{Introduction}





\begin{figure*}[t]
    \centering
    \includegraphics[width=\textwidth]{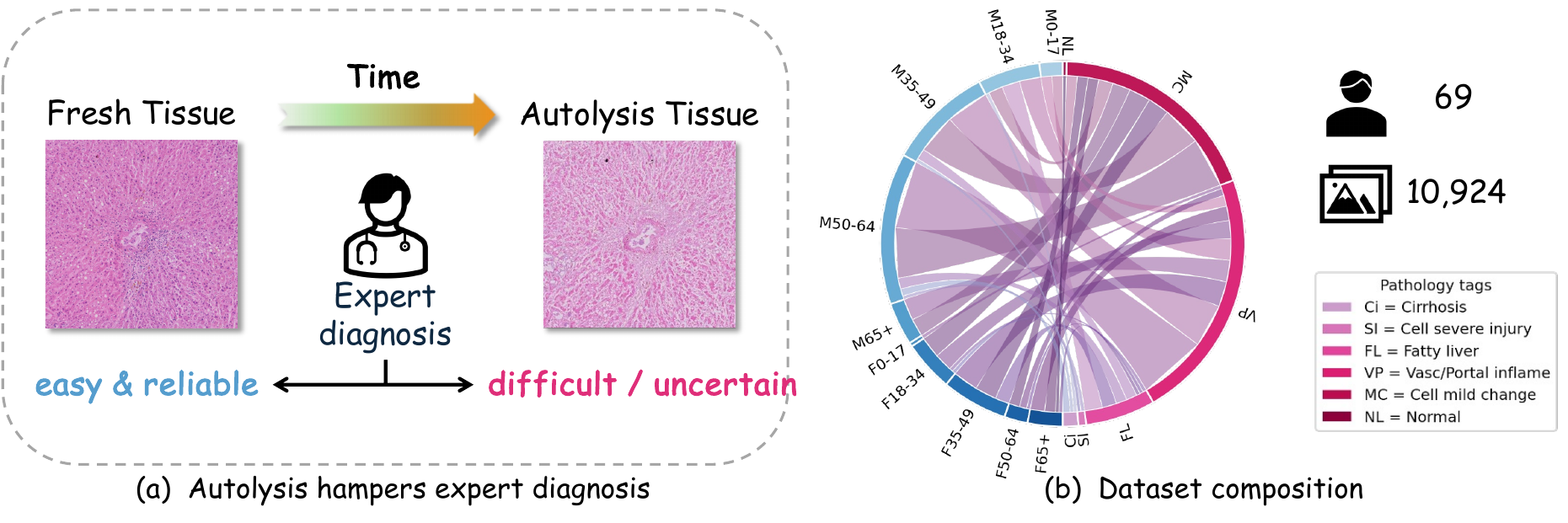}
    \caption{\textbf{Motivation and dataset overview.} (a) As time elapses after exposure, tissue undergoes autolysis that disrupts morphological cues, making expert diagnosis less reliable compared with fresh slides. (b) Composition of our liver pathology dataset, showing the distribution of cases by sex/age groups and their co-occurring pathology tags (chord diagram), together with overall counts of subjects and image patches.}
    \label{fig:fig1}
\end{figure*}

Histopathological examination is the gold standard in forensic medicine for determining cause of death, identifying injury patterns, and diagnosing diseases~\cite{shen2025large,shen2023forensic}. 
However, in contrast to clinical histopathology, postmortem tissues undergo \emph{autolysis}, an irreversible and stochastic degradation driven by enzymatic imbalance following circulatory arrest~\cite{li2017maldi}. This process severely distorts cellular morphology (e.g., nuclear dissolution and tissue disorganization in the liver, as shown in Fig.~\ref{fig:fig1}(a)), introducing substantial diagnostic uncertainty and reliance on subjective expert judgment. 
Consequently, restoring autolyzed histology images toward a plausible pre-autolysis state holds significant potential to improve forensic diagnostic objectivity and reliability.

Although superficially related to image style transfer~\cite{park2020contrastive,cai2024rethinking,zhu2017unpaired,zhao2022egsde} or histological stain normalization~\cite{he2024pst,pati2024accelerating}, forensic autolysis restoration is distinct and more challenging. 
First, autolysis is an irreversible and non-deterministic degradation process, with no symmetric or bijective translation between ``fresh'' and ``autolyzed'' domains. Its progression depends on factors such as temperature, humidity, and postmortem interval, and even under similar conditions the resulting cellular morphology can vary substantially~\cite{madea2023autolysis,secco2025omics,wenzlow2023review}. 
Second, autolysis fundamentally alters tissue morphology rather than merely appearance, violating the structural consistency assumptions common in other translation tasks. 
For instance, in contrast to stain normalization where tissue structures are largely preserved, autolysis causes nuclear dissolution, blurred cell boundaries, and disrupted tissue organization. 
Finally, pixel-wise paired pre-/post-autolysis data are practically unobtainable. 
Together, these characteristics render conventional supervised or cycle-consistent unpaired translation methods ill-suited for this problem.

To address this gap, we formally define the task of \emph{forensic histopathology autolysis restoration}: recovering severely autolyzed images toward plausible pre-autolysis morphologies under unpaired supervision to reveal diagnostically relevant structures. 
We contribute the first dataset, AutoPath, curated by splitting liver specimens into homologous yet unpaired autolyzed and non-autolyzed counterparts, yielding nearly ten thousand images; the cohort composition and pathology-tag distribution are shown in Fig.~\ref{fig:fig1}(b). 
We frame the restoration as a Schr\"odinger Bridge problem~\cite{kim2024unpaired}, learning an optimal stochastic transport between the autolyzed and non-autolyzed domains to handle the large, uncertain morphological changes induced by autolysis.

Critically, we demonstrate that standard image-level metrics (e.g., FID\cite{heusel2017gans}) are misaligned with diagnostic quality in assessing antemortem restoration performance\cite{stein2023exposing,woodland2024feature}: lower FID does not necessarily mean more plausible pathology or higher clinical utility~\cite{sun2023aligning}. We therefore propose a downstream, slide-level diagnostic distribution consistency evaluation grounded in real forensic practice.

To our knowledge, this is the first dedicated study to formulate the task, provide a real-world dataset, and introduce a clinically aligned evaluation framework for autolysis restoration in forensic pathology.
The main contributions of this work are three-fold:
\begin{itemize}
    \item Formal task definition and the first real-world dataset for forensic histopathology autolysis restoration.
    \item A Schrödinger Bridge-based generative framework designed for stochastic, large-magnitude morphological degradation due to autolysis.
    \item A clinically grounded evaluation protocol that highlights the limitations of conventional metrics and aligns assessment with forensic diagnostic practice.
\end{itemize}

\section{Method}

\subsection{AutoPath Data Preparation and Processing}

As shown in Fig.~\ref{fig:pipeline}, we retrospectively collected and screened fresh liver tissue samples from forensic pathology cases, ensuring specimens were free of autolysis at the time of collection. To the best of our knowledge, AutoPath constitutes the first dataset specifically constructed for forensic autolysis restoration. Informed by forensic pathology experts, cellular structures are severely degraded and diagnostically compromised after seven days of ambient exposure. We therefore define specimens exposed for seven days as the \emph{autolyzed} state.

To establish medically meaningful yet unpaired domains, each eligible liver specimen was divided into two spatially adjacent tissue blocks. One block was processed immediately to serve as the \emph{non-autolyzed} sample, while the adjacent block was exposed to ambient conditions for seven days to induce autolysis before slide preparation. This protocol provides the closest possible correspondence in a medical sense, as both slides originate from the same specimen under controlled conditions, despite the fundamental impossibility of obtaining pixel-wise paired samples due to the irreversible nature of histological processing.


In total, samples from 69 patients were collected and processed. All whole-slide images were scanned at 10$\times$ magnification. From these slides, image patches of size $1024 \times 1024$ were randomly sampled. The final dataset consists of 4,962 autolyzed patches and 4,962 non-autolyzed patches for training, with an additional 500 patches per domain reserved as the test set.

\subsection{Task Formulation}
Let $\mathcal{X}$ denote the domain of autolyzed pathological image patches and $\mathcal{Y}$ denote the domain of non-autolyzed patches. Given unpaired samples $\{x_i\}_{i=1}^{N} \sim \mathcal{X}$ and $\{y_j\}_{j=1}^{M} \sim \mathcal{Y}$, the objective of forensic autolysis restoration is to learn a mapping function $T: \mathcal{X} \rightarrow \mathcal{Y}$ that transforms autolyzed images into diagnostically meaningful non-autolyzed representations.

Unlike conventional image-to-image translation tasks, strict pixel-wise or structural correspondence between $\mathcal{X}$ and $\mathcal{Y}$ does not exist. Due to the irreversible biochemical degradation caused by autolysis, significant morphological alterations occur at the cellular level. Consequently, enforcing strong structural consistency constraints between input and output images may contradict the pathological characteristics of the task.

Therefore, the goal of this task is not to preserve exact structural fidelity, but to recover images whose statistical properties and diagnostic distributions are consistent with the non-autolyzed domain.

\subsection{Schrödinger Bridge-based Restoration Framework}
\begin{figure}[t]
    \centering
    \includegraphics[width=\textwidth]{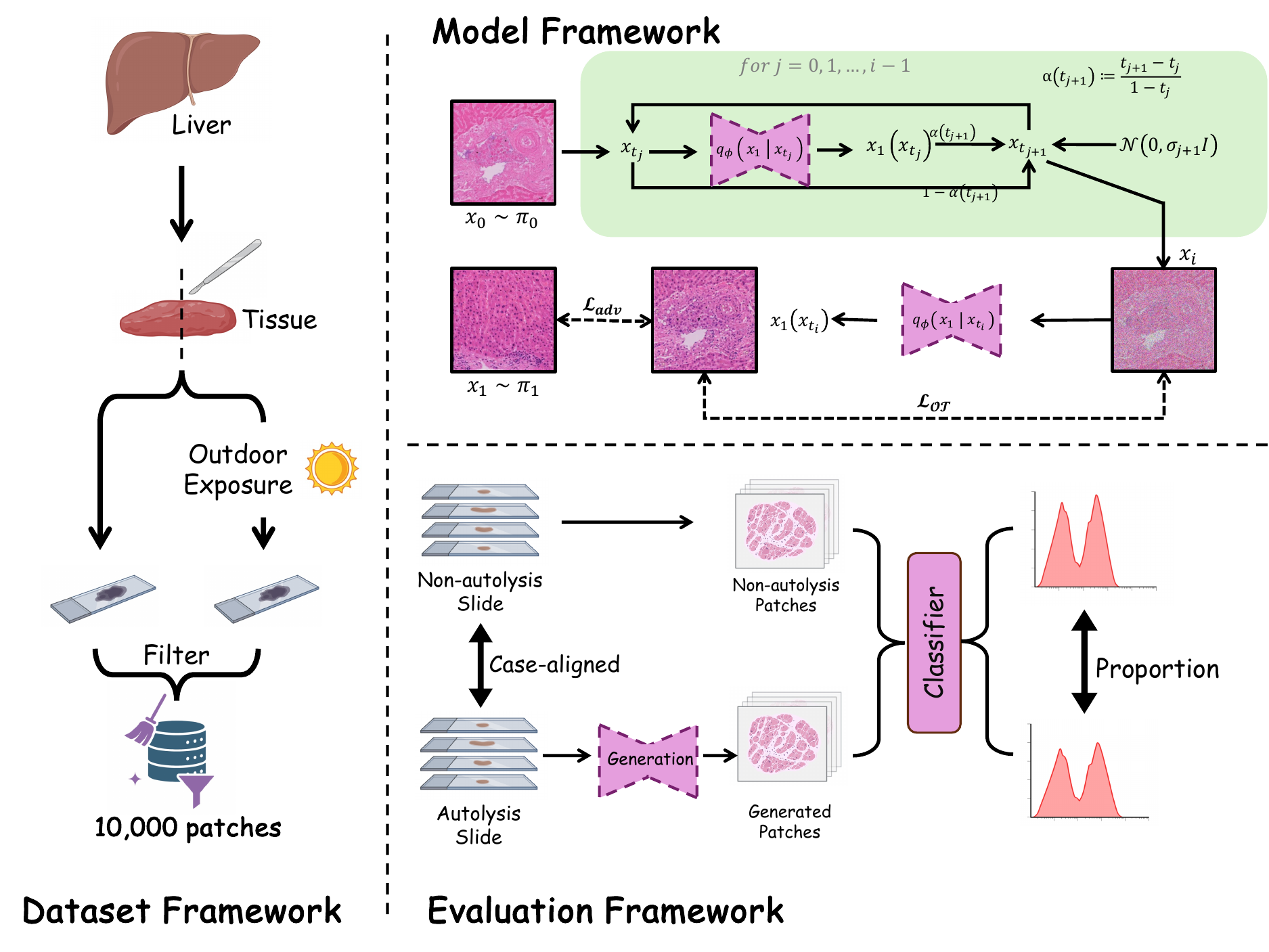}
    \caption{
    Overview of the proposed framework. 
    \textbf{Left: Dataset pipeline.} Fresh liver tissues are divided into adjacent blocks, where one block is immediately processed into a non-autolyzed slide and the other is exposed to ambient conditions for seven days to induce autolysis, followed by patch extraction and filtering. 
    \textbf{Top-right: Schrödinger Bridge-based restoration model.} Autolyzed patches sampled from $\pi_0$ are progressively transported to the non-autolyzed domain $\pi_1$ via multi-step stochastic refinement. 
    \textbf{Bottom-right: Slide-level evaluation framework.} Generated patches and real non-autolyzed patches are classified to obtain diagnostic category distributions, and slide-level proportion consistency is used to assess diagnostic validity.
    }
    \label{fig:pipeline}
\end{figure}

As shown in Fig.~\ref{fig:pipeline}, we formulate the restoration of autolyzed forensic histopathology images as a Schrödinger Bridge (SB) problem, aiming to find the optimal stochastic transport between the empirical distribution of $\pi_0$ and the non-autolyzed distribution $\pi_1$. 
The SB identifies a stochastic process $Q$ that minimizes the Kullback-Leibler (KL) divergence to a reference Wiener process $W_\tau$, under the boundary constraints that $Q_0 = \pi_0$ and $Q_1 = \pi_1$:
\begin{equation}
Q^{SB} = \arg\min_{Q} 
D_{KL}(Q \| W_\tau)
\quad
\text{s.t. } Q_0 = \pi_0, \; Q_1 = \pi_1,
\end{equation}
where $\tau$ controls the diffusion variance. We adopt a discrete-time formulation, dividing the time interval $[0,1]$ into steps $\{t_i\}_{i=0}^N$. A time-conditional generator $q_\phi(x_1|x_{t_i}, t_i)$ is trained to predict target-domain (non-autolyzed) samples from intermediate states. The overall training objective combines an adversarial alignment term ($\mathcal{L}_{Adv}$) for distribution alignment and the Schrödinger Bridge transport loss ($\mathcal{L}_{SB}$):
\begin{equation}
\mathcal{L}
=
\mathcal{L}_{Adv}
+
\lambda_{SB}\mathcal{L}_{SB}.
\end{equation}
The adversarial term $\mathcal{L}_{Adv}$ enforces marginal alignment by matching generated target samples $\hat{x}_1 \sim q_\phi(x_1|x_{t_i},t_i)$ to real non-autolyzed patches $x_1 \sim \pi_1$ via a discriminator, without requiring any paired correspondence. 
In contrast, the Schr\"odinger Bridge transport loss $\mathcal{L}_{SB}$ shapes the intermediate-to-target dynamics by encouraging low-cost stochastic transport under the diffusion reference: it is implemented as an entropic transport objective that combines a quadratic transport cost between $x_{t_i}$ and $\hat{x}_1$ with an entropy regularizer to avoid degenerate couplings and to reflect the stochastic nature of autolysis. 
Together, $\mathcal{L}_{SB}$ guides the bridge evolution, while $\mathcal{L}_{Adv}$ anchors the final outputs to the target-domain distribution.

Crucially, we remove structural regularization terms commonly used in image translation. Since autolysis causes significant and irreversible morphological degradation, enforcing structural consistency between input and output may preserve autolytic artifacts rather than facilitate genuine restoration. Our model therefore focuses exclusively on learning the distribution-level mapping from $\pi_0$ to $\pi_1$, without imposing pixel- or patch-level correspondence.

During sampling, given a predicted target sample $x_1(x_{t_i}) \sim q_\phi(x_1|x_{t_i})$, the state is updated as:
\begin{equation}
x_{t_{i+1}}
=
\alpha_{i+1} x_1(x_{t_i})
+
(1 - \alpha_{i+1}) x_{t_i}
+
\epsilon,
\end{equation}
where 
$\alpha_{i+1}=\frac{t_{i+1}-t_i}{1-t_i}, \, \epsilon \sim \mathcal{N}(0, \alpha_{i+1}(1-\alpha_{i+1})\tau I)$.

The model is trained by randomly sampling time steps and jointly optimizing $\mathcal{L}_{Adv}$ and $\mathcal{L}_{SB}$.
At inference, an autolyzed input image is iteratively refined through the discretized bridge, yielding the final restored image that resides in the non-autolyzed distribution $\pi_1$.


\section{Experiments}

\subsection{Experimental Setup}

All experiments are conducted on a server equipped with an NVIDIA RTX 4090 GPU for both training and inference. We perform experiments on the AutoPath dataset, which is specifically constructed for forensic autolysis restoration. To comprehensively evaluate restoration quality, we adopt a three-part evaluation protocol. First, we report image-level generative metrics, including the Fréchet Inception Distance (FID), to assess global distribution similarity. Second, we conduct expert-based human validation to evaluate morphological plausibility from a forensic pathology perspective. Third, we perform slide-level diagnostic distribution consistency analysis to measure case-level diagnostic alignment between restored slides and their corresponding non-autolyzed references.

\subsection{Antemortem Image Restoration Results}

We evaluate the proposed method on the AutoPath dataset under the experimental setup described above. In addition to our approach, we compare against representative unpaired image translation methods, including CycleGAN, CUT, DenseNorm (DN), and KIN. All models are trained and tested under identical data splits for fair comparison.

\begin{table*}[t]
\centering
\caption{\textbf{Quantitative and slide-level evaluation on AutoPath.}
We report patch-level FID (lower is better) with expert-based validation (higher is better), and slide-level diagnostic distribution consistency on 69 cases. Higher is better for WSDC and High-Conf Ratio; lower is better for WKL.}
\label{tab:autopath_merged}
\setlength{\tabcolsep}{7pt}
\renewcommand{\arraystretch}{1.12}
\begin{tabular}{l cc ccc}
\hline
\multirow{2}{*}{\textbf{Method}}
& \multicolumn{2}{c}{\textbf{Patch-level}}
& \multicolumn{3}{c}{\textbf{Slide-level (69 cases)}} \\
\cline{2-6}
& \textbf{FID} $\downarrow$
& \textbf{Expert Val.} $\uparrow$
& \textbf{WSDC} $\uparrow$
& \textbf{WKL} $\downarrow$
& \textbf{High-Conf} $\uparrow$ \\
\hline
CycleGAN
& 51.24 & 18.00\%
& 0.748 & 0.406 & 0.885 \\

DN
& 51.87 & 2.38\%
& 0.723 & 0.378 & 0.840 \\

KIN
& \textbf{39.78} & 18.75\%
& 0.740 & 0.367 & 0.830 \\

CUT
& 53.46 & 17.65\%
& 0.769 & 0.399 & 0.889 \\

\textbf{Ours}
& 55.63 & \textbf{23.81\%}
& \textbf{0.799} & \textbf{0.343} & \textbf{0.897} \\
\hline
\end{tabular}
\end{table*}

Quantitative image-level results measured by FID are reported in Table~\ref{tab:autopath_merged}. Our method does not achieve the lowest FID score among the compared approaches.

To further assess morphological realism from a forensic perspective, we conduct expert-based human validation. For each evaluation instance, we randomly sample one restored patch from each method and additionally sample one real non-autolyzed patch as reference. These images are presented in a blinded manner to three board-certified forensic pathologists, who are asked to select the image that appears most realistic and diagnostically consistent with non-autolyzed morphology. The reported percentages in Table~\ref{tab:autopath_merged} represent the average selection ratio across the three experts.

\begin{figure}[t]
    \centering
    \includegraphics[width=0.96\textwidth]{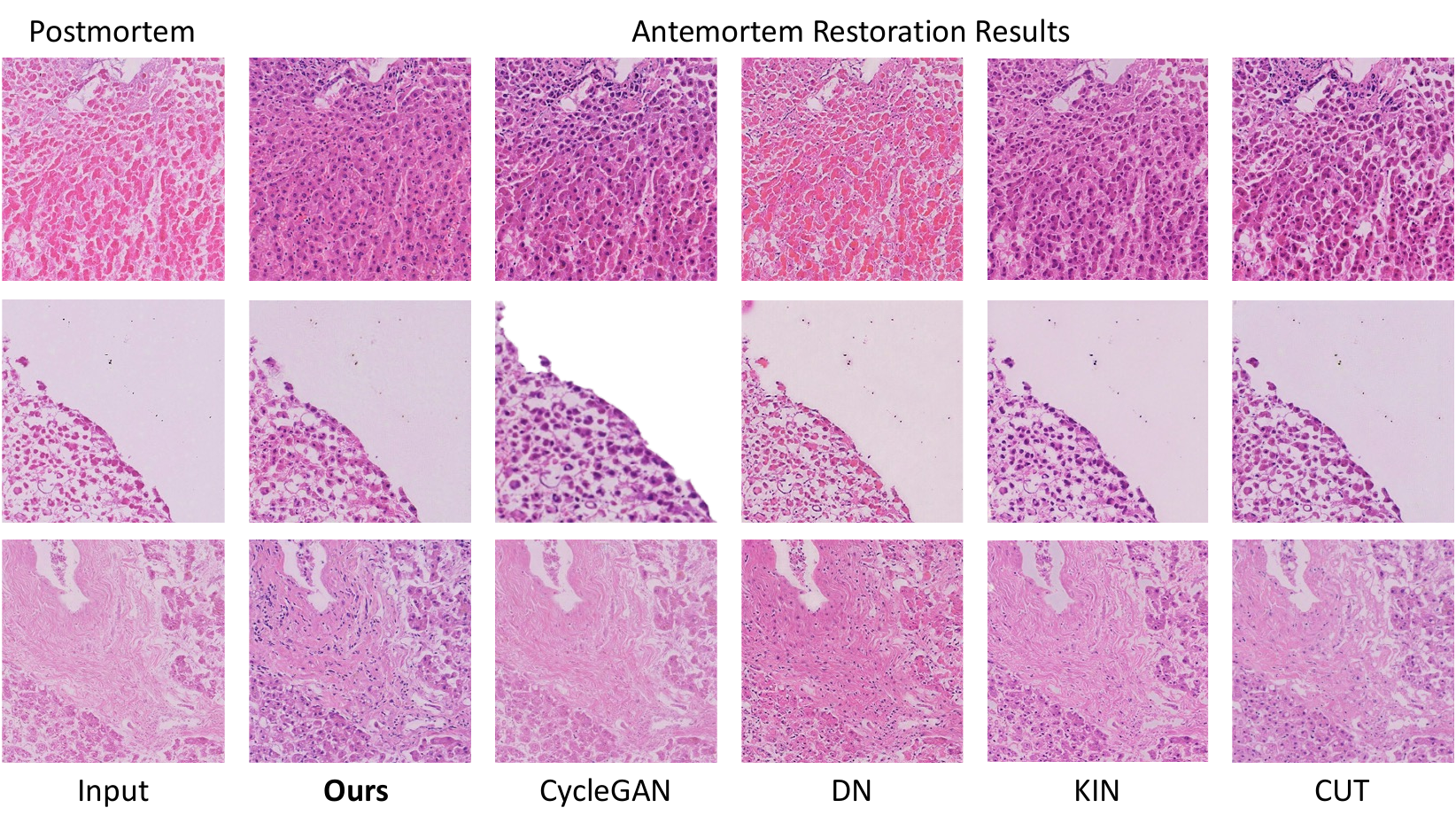}
    \caption{
    Qualitative comparison of autolysis restoration on AutoPath. We visualize representative autolyzed liver patches and the corresponding restoration results produced by different methods. Existing approaches often retain or amplify autolysis-induced artifacts (e.g., blurred nuclei and disrupted tissue organization), whereas Schrödinger Bridge-based model yields more plausible cellular morphology and tissue architecture according to forensic pathology experts.
    }
    \label{fig:visual}
\end{figure}

Despite having a higher FID score, our method achieves the highest expert preference (23.81\%). This trend is also supported by qualitative comparisons in Fig.~\ref{fig:visual}, where competing methods frequently retain autolysis-induced degradation patterns (e.g., blurred nuclei and disrupted tissue organization) or introduce over-smoothed textures, while our Schr\"odinger Bridge-based model produces more plausible cellular morphology and tissue architecture that better resemble non-autolyzed references.

In contrast, other competing methods with lower FID values often preserve fine-grained morphological traces introduced by autolysis, which may inadvertently retain degradation artifacts. This discrepancy suggests that image-level generative metrics emphasize feature similarity and structural fidelity, which do not necessarily correspond to diagnostic plausibility in forensic pathology.

\subsection{Slide-level Diagnostic Distribution Consistency Evaluation}

While patch-level evaluation provides local visual assessment, forensic diagnosis is ultimately performed at the case level. Therefore, we conduct slide-level validation on all 69 cases in the AutoPath dataset to assess whether restored images preserve overall diagnostic consistency. For each case, autolyzed and non-autolyzed slides are prepared from spatially adjacent tissue regions during specimen processing, ensuring comparable pathological contexts. Under this assumption, case-aligned autolyzed and non-autolyzed slides are expected to exhibit similar diagnostic category distributions. A video illustration of the specimen preparation procedure is provided in the \textbf{supplementary material} for reference.

All autolyzed patches are first restored using different methods. A classifier, built upon a CONCH-based feature extractor~\cite{lu2024visual} and fine-tuned with approximately 300 manually annotated patches by forensic pathology experts, is then applied to \emph{all} patches from all slides (on the order of $5\times10^{4}$ patches in total) for both restored and real non-autolyzed images.  Patch-level predictions are aggregated to obtain slide-level diagnostic distributions. We quantify distributional agreement using \emph{Weighted Slide-level Diagnostic Consistency} (WSDC) and \emph{weighted KL divergence} (WKL), where higher WSDC and lower WKL indicate closer alignment with the case-aligned non-autolyzed reference. In addition, we report the \emph{High-Confidence Ratio}, measuring the proportion of predictions above a confidence threshold, to reflect the stability of diagnostic decisions.

Quantitative results are summarized in Table~\ref{tab:autopath_merged}. Our method achieves the highest WSDC (0.799) and the lowest WKL (0.343), indicating the closest alignment with case-aligned non-autolyzed references. Moreover, it attains the highest High-Confidence Ratio (0.897), suggesting more stable and reliable diagnostic behavior at the case level.

In contrast, baseline methods exhibit lower weighted consistency and larger weighted divergence, despite some achieving competitive patch-level FID scores. These findings highlight that slide-level diagnostic distribution consistency provides a more practically meaningful evaluation criterion for forensic autolysis restoration than conventional patch-level generative metrics.





\section{Conclusion}

We present the first systematic study of forensic histopathology autolysis restoration, a challenging task defined by irreversible, stochastic morphological degradation under unpaired supervision. To support this investigation, we introduce AutoPath, the first homologous yet unpaired dataset curated specifically for this problem. Moving beyond conventional image-level metrics, we propose a forensically grounded evaluation framework that incorporates both expert-level patch consistency and case-level diagnostic distribution consistency. Our experiments demonstrate that widely adopted generative metrics (e.g., FID) do not adequately reflect diagnostic realism in this context, whereas distribution-level consistency offers a more meaningful measure of forensic applicability. Overall, this work establishes autolysis restoration as a valuable and underexplored direction in computational pathology and underscores the need for task-driven evaluation standards in pathological image restoration.

\begin{credits}
\subsubsection{\ackname} This work was supported in part by the National Natural Science Foundation of China (Nos. T2522028 and 12326616), Natural Science Basic Research Program of Shaanxi (No. 2024JC-TBZC-09), and Shaanxi Provincial Key Industrial Innovation Chain Project (No. 2024SF-ZDCYL-02-10).

\subsubsection{\discintname}
The authors have no competing interests to declare that are relevant to the content of this article.
\end{credits}

%
%
%
\bibliographystyle{splncs04}
\bibliography{mybibliography}
%




\end{document}